\documentclass{article}

\usepackage{arxiv}

\usepackage{url}

 \usepackage{multirow}
\usepackage{amsmath}
\usepackage{tikz}
\usetikzlibrary{shapes.geometric, arrows}
\tikzstyle{startstop} = [rectangle, rounded corners, minimum width=1cm,  minimum height=0.8cm,text centered, text width=1.15cm, draw=white, fill=gray!60]
\tikzstyle{arrow} = [thick,->,>=stealth]
\tikzstyle{process} = [rectangle, rounded corners, minimum width=1cm, minimum height=0.8cm, text width=1.15cm, text centered, draw=black, fill=white!30]
\usepackage{subcaption}
\usepackage{tikz}
\def\checkmark{\tikz\fill[scale=0.4](0,.35) -- (.25,0) -- (1,.7) -- (.25,.15) -- cycle;} 
\newcommand\copyrighttext{%
  \footnotesize \textcopyright 2020 IEEE. Personal use of this material is permitted.
  Permission from IEEE must be obtained for all other uses, in any current or future 
  media, including reprinting/republishing this material for advertising or promotional 
  purposes, creating new collective works, for resale or redistribution to servers or 
  lists, or reuse of any copyrighted component of this work in other works. \\
This paper was accepted for publication in the 25th International Conference on Pattern Recognition (ICPR2020). 
 }
\newcommand\copyrightnotice{%
\begin{tikzpicture}[remember picture,overlay]
\node[anchor=south,yshift=10pt] at (current page.south) {\fbox{\parbox{\dimexpr\textwidth-\fboxsep-\fboxrule\relax}{\copyrighttext}}};
\end{tikzpicture}%
}
\begin{document}
\copyrightnotice
%
\title{Detecting Anomalies from Video-Sequences: \\a Novel Descriptor}

\author{Giulia Orrù, Davide Ghiani, Maura Pintor, Gian Luca Marcialis, Fabio Roli\\
Department of Electrical and Electronic Engineering\\ University of Cagliari, Italy\\
{\tt\small \{giulia.orru, maura.pintor, marcialis, roli\}@unica.it, davideghiani@gmail.com}}

\maketitle

\begin{abstract}
We present a novel descriptor for crowd behavior analysis and anomaly detection. The goal is to measure by appropriate patterns the speed of formation and disintegration of groups in the crowd. This descriptor is inspired by the concept of one-dimensional local binary patterns: in our case, such patterns depend on the number of group observed in a time window. An appropriate measurement unit, named ``trit'' (trinary digit), represents three possible dynamic states of groups on a certain frame. Our hypothesis is that abrupt variations of the groups' number may be due to an anomalous event that can be accordingly detected, by translating these variations on temporal trit-based sequence of strings which are significantly different from the one describing the ``no-anomaly'' one. Due to the peculiarity of the rationale behind this work, relying on the number of groups, three different methods of people group’s extraction are compared. Experiments are carried out on the Motion-Emotion benchmark data set. Reported results point out in which cases the trit-based measurement of group dynamics allows us to detect the anomaly. Besides the promising performance of our approach, we show how it is correlated with the anomaly typology and the camera's perspective to the crowd's flow (frontal, lateral). \end{abstract}

\section{Introduction}
Anomalous event detection in crowded environments \cite{DBLP:books/el/MCSS2017} by computer vision (CV) techniques is an investigation field with a strong impact in the physical security domain. In this context, the repressive action of the police could be helped if an automatic mean of fast search of anomalous events over a video-sequence would be available. Moreover, an automated system could give significant help if it would be able to predict the occurrence of such an event in real-time. 

Although the notable efforts in terms of anomalous events detection and classification, especially by using deep networks \cite{cnncrowd}, experimental results showed these systems cannot yet be applied in real environments without knowing their limitations. 

The main motivation is that it is not easy to define in a mathematical closed-form what an anomalous event is \cite{DBLP:books/el/MCSS2017}. Each application has its own specificities which may be associated with behavioural anomalies. Accordingly, behavioural biometrics \cite{Vishwakarma2012ASO}, for example, the gestural analysis or the facial expression evaluation, are useful to assess a certain class of anomaly, such as physical or verbal violence \cite{Sultani_2018_CVPR}.

In this paper, we focused on the assessment of how much a ``state variable'' denoting an anomalous event is far from the correspondent value of the ``quiet'' state, which denotes the absence of anomaly, over time \cite{6898845}. We took into account that often video-surveillance cameras are very far from the foreground scene and give very few insights about the gestures, facial expressions and actions of individuals, although these are potentially evidence of anomalous events.

In this paper, we considered that crowds tend to assume an aggregative or dispersive behavior whose rapidity depends on the nature of the occurring event. In particular, rapid and random aggregative or dispersive behavior occurs in correspondence of an anomaly: this is evident in events causing the crowd's panic. The rapidity modelling is, therefore, the core of our approach; in other words, our goal is to assess the rapidity of the crowd's answer to a certain external stimulus.

We designed a descriptor based on accounting the number of people groups detected into the scene, and how it increases or decreases over time. A descriptor that works accordingly allows us to “explain” what is happening by evaluating how much it changes. Nevertheless, it may become less effective when a normal transition in the groups' number tends to be too similar to that supposed for an anomalous event, thus causing false alarms, even if they could be explained in retrospect. This explanation ability is still difficult to be obtained by a black-box approach like a deep network-based one \cite{escalante:hal-01991623}.

Two modules compose our system. The first one is the clustering module whose aim is to detect the people groups in the scene – we implemented several state-of-the-art algorithms \cite{santoro2010crowd,viola2001rapid,lienhart2002extended,opencv_library} beside the manual counting of the groups. The second one is the original contribution of this paper: a novel temporal descriptor aimed to represent the variations of the group numbers over time.

The one-dimensional local binary pattern \cite{chatlani2010local} inspired our descriptor. By a time-window sliding over a certain video-sequence's set of frames, and centred on a certain temporal instant, we compare the number of groups computed at that instant with the groups count after and before it. Three cases may happen: the number of groups increases, decreases, or remains unaltered. This leads to have, for each time-window, a set of ``trinary'' states (increase/decrease/unaltered), that we called string of ``trits'' (trit=\textbf{\textit{tri}}nary digi\textbf{\textit{t}}). After collecting a certain number of strings, we evaluated the histograms of the trinary states occurrences, thus obtaining a temporal description of the groups' dynamic over that set of frames. The deviation of each histogram from the “quiet” state-related histogram allows us to decide if an alarm must be output. 

We adopted the Motion-Emotion data set \cite{rabiee2016novel} to benchmark our system, and, where possible, we compared that with other state-of-the-art methods. Unfortunately, the lack of a rigorous experimental protocol shared by all research groups limits the significance of this comparison, since the same data set is used for different goals. Anyway, we computed the performance parameters usually adopted in terms of the so-called F1 score and precision/recall. Beside them, we discussed the pros and cons of the system by correlating the system's behaviour with the video-sequence. This allowed us to observe the change in the trit-based state histogram according to what happens over the video. In other words, we verified if our hypothesis is supported as well as the impact of the noise introduced by non-zero-detection-error modules (in particular, the group counting module).

Although the obtained performance is not yet ready for practical use of the system, reported experiments are in agreement with our claim, that is, we propose a novel and explainable method to anomaly detection in mass-based video-surveillance applications. This method relies on a hypothesis physically verifiable, thus it acts as full white-box whose parameters are correlated to the specific detection goal.

The paper is organized as follows. Section 2 makes the point about the state-of-the-art on modeling the crowd behavior. Section 3 describes the proposed system. Experimental results are reported in Section 4, and the concluding Section is a frank discussion about the pros and cons of this proposal.

\section{Anomalous events and previous works}
\begin{figure}[ht]
\centering
\includegraphics[width=.23\textwidth]{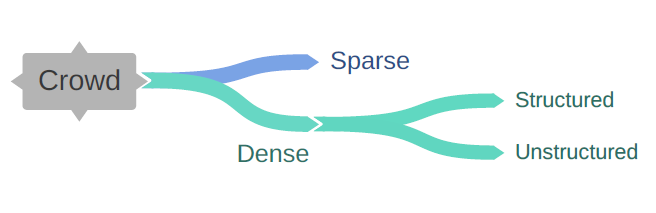}
\caption{Taxonomy of crowds.}
\label{fig:crowd}
\end{figure}

The crowd analysis is the study of the natural movements of people and objects in the scene. 
Such a group of people is often named ``crowd''. The crowd analysis  \cite{6898845} can be conducted at the macroscopic level by considering the crowd global motions, or at the microscopic level that is focused on the movements of each individual.
Based on the density of the crowd in the scene, crowd analysis methods can be grouped into two major categories, sparse crowd analysis and dense crowd analysis (Fig. \ref{fig:crowd}) \cite{10.1007/978-981-10-3770-2_3}. The density of a crowd is commonly an estimate of how many people are in a delimited area and it concerns the possibility of distinguishing the individuals and the background. Therefore, \textit{sparse crowd} is a crowd where the individuals are separable and identifiable and occlusions are limited, while in the case of a \textit{dense crowd}, people are crammed together and it is impossible to distinguish the shape of each individual due to the high level of occlusions.

\begin{figure}[ht]
\centering
\includegraphics[width=.45\textwidth]{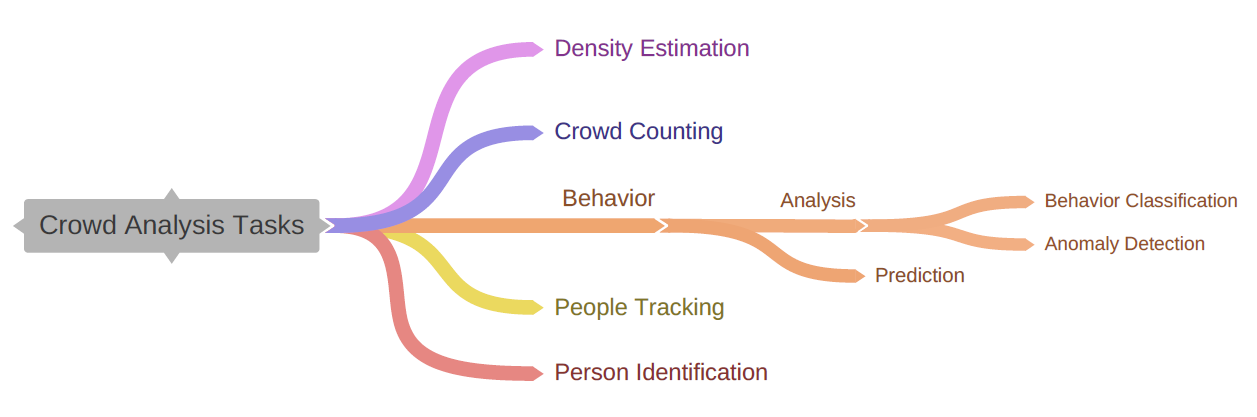}
\caption{Taxonomy of crowd analysis task.}
\label{fig:crowdtask}
\end{figure}

The dense crowds can be divided into structured and unstructured crowds according to the motion present in the scene. If the crowd moves coherently on a macroscopic level with a common and constant direction, the crowd is structured. If the crowd is characterized by chaotic or random movements and at a microscopic level from different directions and different gestures, the crowd is called ``unstructured''. 
Crowd analysis allows to face a large number of tasks (Fig. \ref{fig:crowdtask}) which can be schematized by macro areas as density estimation \cite{wu2006crowd}, crowd counting \cite{chan2008privacy}, behavioral analysis and prediction \cite{mehran2009abnormal}, people tracking \cite{rodriguez2009tracking} and person identification.

Based on the task, the crowd analysis involves extremely different methods and disciplines. The analysis is performed in different levels, from raw to elaborate features.

\begin{figure}[ht!]
\centering
\includegraphics[width=0.35\textwidth]{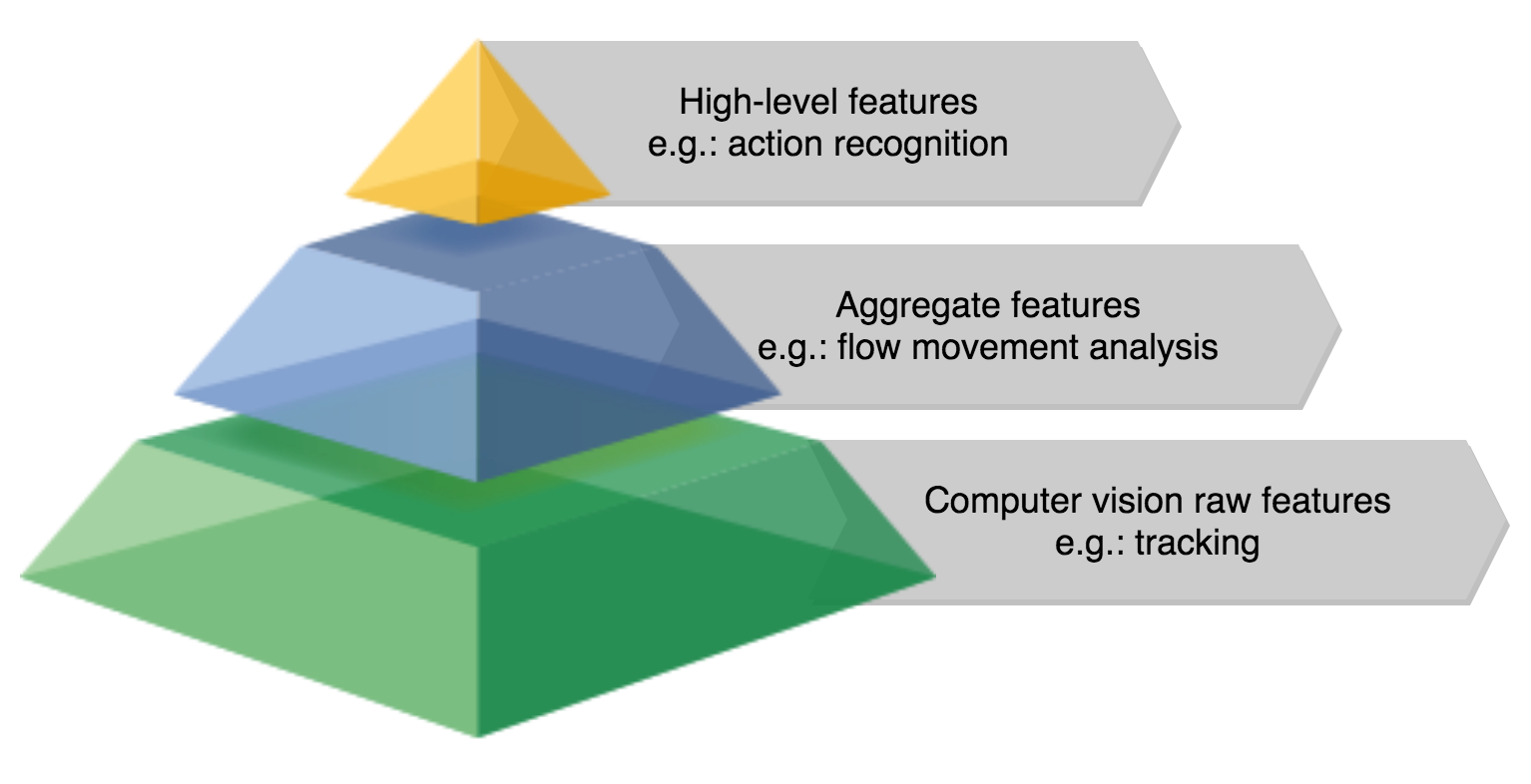}
\caption{Overview of features used for CV-based crowd analysis.}\label{fig:pyramid}
\end{figure}

Figure \ref{fig:pyramid} shows an overview of the features used for computer vision-based crowd analysis. Usually a bottom-up approach is adopted, starting from a voluminous amount of low-level features and refining the information through processing and selection, up to high-level features \cite{DBLP:books/el/MCSS2017}. 

At the base, a large set of \textbf{raw features} is collected. The raw features are extracted from the sensors, such as one or more videocameras. Raw features, such as the RGB level of each pixels, can be used, but the extraction of specific features is often done, such as for the optical flow-based features \cite{horn1981determining} from the frames sequence or the textural-based features \cite{hao2017extracting} from the single frame. 

At the middle level, \textbf{aggregate features} are obtained from the raw features, usually by reducing the dimensionality and number of features. At this level, individuals can be detected and tracked, or groups of people can be identified \cite{7583749} by the extracted features. However, it is not possible to have the clear understanding of the scenario, since they are still lacking of the relationships with the whole context. 

The \textbf{high-level features} allow to relate the previous features sets with their context. This step includes psychological and social knowledge in order to understand the targeted event. The features processing is strongly affected by the accuracy of the previous ones, and requires context-dependent modelling. The final interpretation of the scene as given by the third-level features is left to the humans.

In the anomaly detection task, numerous descriptors are based on raw or aggregate features, generally called low-level features, as Mixture of Dynamic Texture (MDT) \cite{mahadevan2010anomaly} and the spatio-temporal descriptors Histogram of Gradient (HOG) \cite{6619181} and Histogram of Optical Flow (HOF) \cite{10.1007/978-3-319-26561-2_49}.
Others combine motion information and context information, as Ref. \cite{article} which extracts the local binary patterns for obtaining contextual information.
In recent years, deep learning-based anomaly detection algorithms have become increasingly popular and allowed to achieve very accurate results.

Despite their high accuracy, the black-box nature of deep-learning approaches makes these methods poorly interpretable \cite{Sydney2019DeepLF} and vulnerable to adversarial attacks \cite{adv}. In this work we propose a time descriptor of the crowd that allows to give an explanation of the detection of the anomaly.

\section{A novel descriptor for anomalous events detection in high-density crowds}

\subsection{Crowd model}

\begin{figure}[ht]\centering

\begin{subfigure}{3.5cm}
    \centering\includegraphics[width=3.5cm]{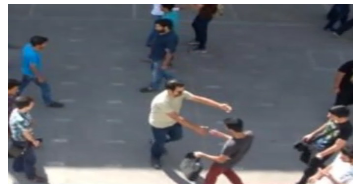}
    \caption{}
  \end{subfigure}
  \begin{subfigure}{3.5cm}
    \centering\includegraphics[width=3.5cm]{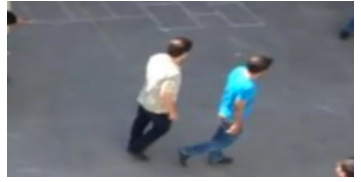}
    \caption{}
  \end{subfigure}
 
  \begin{subfigure}{3.5cm}
    \centering\includegraphics[width=3.5cm]{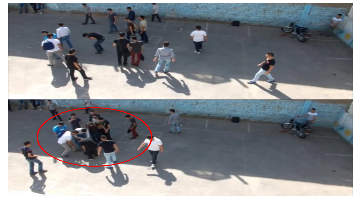}
    \caption{}
  \end{subfigure}
  \begin{subfigure}{3.5cm}
    \centering\includegraphics[width=3.5cm]{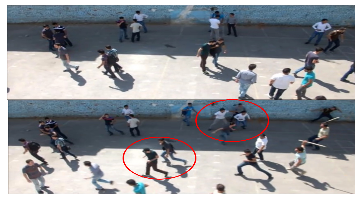}
    \caption{}
  \end{subfigure}
\caption{Examples of crowd dynamics: (a) slow formation; (b) slow breaking-up; (c) fast formation; (d) fast breaking-up.} \label{fig:example-formation-and-splitting}
\end{figure}

Crowd behaviour is strictly related with crowd motion patterns. In a sparse crowd scenario, these patterns are defined by individuals that move in the scene, and eventually form or break-up groups. We identify a group as a set of individuals that: 1) move together, speak together, or perform a common action; 2) are physically close; 3) are directed in a common centre of focus.
Our system is based on the hypothesis that anomalous events happen when multiple \textit{group formation events} and \textit{group breaking-up events} suddenly appear in the scenario. It is worth noting that these events happen likewise in absence of anomaly, but with different dynamics. Figure \ref{fig:example-formation-and-splitting} shows examples of slow and fast crowd dynamics. By counting the number of groups in the scene and monitoring its variation over time, it is possible to detect such variations and use them for the purpose of anomaly detection in crowd behaviour.

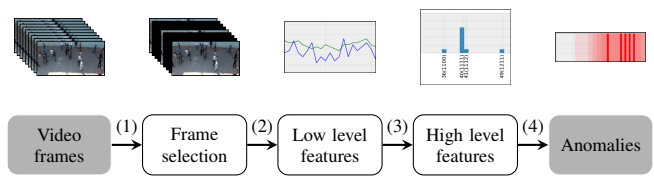
\begin{figure}[ht]\centering
\begin{tikzpicture}[node distance=1.8cm]
\tikzstyle{every node}=[font=\scriptsize]

\node (frames) at (0,0) {\includegraphics[width=1.2cm]{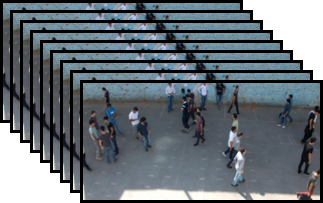}};
\node (selection) [right of=frames] {\includegraphics[width=1.2cm]{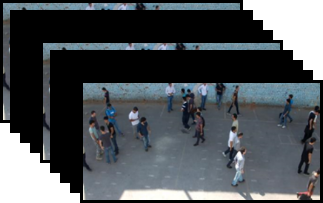}};
\node (counts) [right of=selection] {\includegraphics[width=1.2cm]{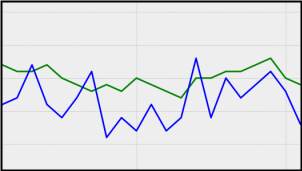}};
\node (hist) [right of=counts] {\includegraphics[width=1.2cm]{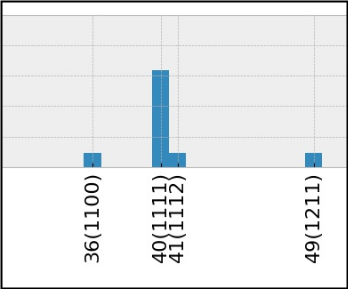}};
\node (anomalies) [right of=hist] {\includegraphics[width=1.2cm]{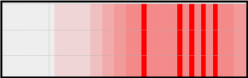}};

\node (framesd) [startstop, below of=frames, yshift=0.5cm] {Video frames};
\node (selectiond) [process, below of=selection, yshift=0.5cm] {Frame selection};
\node (countsd) [process, below of=counts, yshift=0.5cm] {Low level features};
\node (histd) [process, below of=hist, yshift=0.5cm] {High level features};
\node (anomaliesd) [startstop, below of=anomalies, yshift=0.5cm] {Anomalies};

\draw [arrow] (framesd) -- node[anchor=south]{(1)} (selectiond);
\draw [arrow] (selectiond) -- node[anchor=south]{(2)} (countsd);
\draw [arrow] (countsd) -- node[anchor=south]{(3)} (histd);
\draw [arrow] (histd) -- node[anchor=south]{(4)} (anomaliesd);

\end{tikzpicture}
\caption{Anomaly detection pipeline. (1) A subset of the total frames is selected from the whole sequence of frames; (2) Low level features are extracted for obtaining the number of groups in each scene; (3) High level features compute statistics of dynamics patterns; (4) Anomalies are obtained through thresholding a specific pattern.} \label{fig:pipeline}
\end{figure}

Our system consists of three layers, shown in Figure \ref{fig:pipeline}. First, we select a subset of the total number of frames for avoiding a dense analysis (1). In the second layer, we count the number of groups in the scene for each selected frame with computer vision techniques (2). This will serve as low-level features and holds little information about the anomalies. Finally, the output of this block passes through a refinement layer (3) that describes with a histogram each analysed sequence and links the low-level features with our crowd model. A threshold is applied to the central bin of the histogram, characterizing the state of quiet, and acts as a trigger for detecting the anomaly. 


\subsection{Low level features}
\label{sec:lowlevfeat}
The input of our system is a video sequence of $N_{tot}$ frames. We select one frame each $F$, the number of frames that we skip between two subsequent group counts, obtaining $\frac{N_{tot}}{F}$ frames. 
Low level features are intended to count the number of groups in the scene along time. We compare three automatic methods for counting the groups, counting each $F$ frames and the manual counting as ground truth.

\noindent \textbf{Clustering of optical flow (COF).} We use dense optical flow (OF)\cite{horn1981determining} for tracking moving blocks of pixels from one (selected) frame to another. Then, a Density-Based Spatial Clustering of Applications with Noise (DBSCAN) algorithm is used for grouping into blocks the optical flow vectors that: 1) have the same angle of movement (phase of the OF); 2) have the same speed (magnitude of the OF); 3) are spatially close to each other (x,y coordinates of the pixel). We use the clustering algorithm for obtaining the number of clusters, rather than assigning each point to a cluster.

\noindent \textbf{Cascade detector (CD).} We estimate the number of groups by counting the number of bounding boxes resulting from the application of an off-the-shelf Multi Scale Object Detector implemented in OpenCV \cite{opencv_library}. 
This method detects individual persons rather than clusters, but our hypothesis is that for each group only the front person will be completely visible. We use the number of bounding boxes as a rough estimation of the number of groups. This method is the fastest, but highly inaccurate in the absolute count.

\noindent \textbf{Blob detector (BD).} We implemented our own version of a bounding box detector. First, it removes the background and applies gaussian blurring to the image in order to reduce the noise. After, it binarizes the image and highlights the shapes of the blobs, i.e., the groups that we are willing to count. Finally, we apply dilation and opening for extracting the contour of the shapes, then we count the number of detected contours and this will act as our count.

To the bounding box detection is not applied any post-processing.
The output of this block is a raw count of groups for each selected frame. It is important to mark that we are not interested in the accurate value of this count, rather a correct detection of the abrupt changes. This means that the group counting algorithm is supposed to capture a trend rather than exact counts.

\subsection{High level features}
\begin{figure}[ht]
    \centering
    \includegraphics[width=11cm]{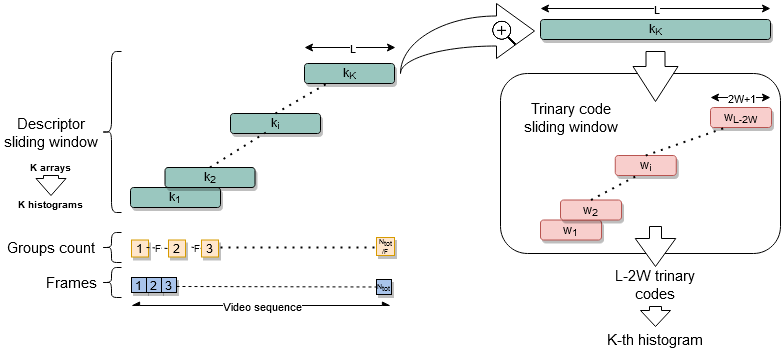}
    \caption{High-level features extraction scheme. The group counts are divided into $K$ arrays of size $L$ through a first sliding windows. Each array is further divided with another sliding window in $L-2W$ subgroups of size $2W-1$, each of which generates a trinary code. The trinary codes transformed into decimal values are collected in a histogram.}
    \label{fig:highlev}
\end{figure}

Once the group count has been extracted, we can proceed with the high-level feature extraction (Fig. \ref{fig:highlev}). We designed a descriptor that can be applied to any one-dimensional signal for anomaly detection. The output of this layer is a sequence of labels that marks where the anomaly starts in each video.

After the first layer of feature extraction, using a sliding window of size $L$, we obtain $K$ arrays. Since scrolling the window on the $\frac{N_{tot}}{F}$ selected frames of one element at a time, $K=\frac{N_{tot}}{F} - L +1 $. For each array $k_i$, with $i=1,...,K$, we apply the following algorithm: 
\begin{enumerate}
    \item We scroll the array $k_i$ with another sliding window of size $2W + 1$,  being $W$ a positive integer smaller than half of $L$, getting $L-2W$ windows. For each of these windows, we compare the number of groups of the central value $c$ with the surrounding ones. By the comparison of each element $c_{j}$ of the sliding window, being $j=0,..,2W$ with the central value $c$, we obtain a trinary code by applying Eq. \ref{eq:trinary}.

\begin{equation}
\small{
t_j = 
\begin{cases}
0 \text{\quad} if \text{\quad}  c_j-c >T \\
1 \text{\quad} if \text{\quad} | c-c_j| <= T \\
2 \text{\quad} if \text{\quad} c-c_j  > T\\
\end{cases}
}
\end{equation}{\label{eq:trinary}}

The descriptor threshold T is set to reduce the noise influence in the count. This allows the state of quiet to evolve over time: if the groups count gradually increases or descends the sequence will not be identified as anomalous. We drop the central value $c$ for avoiding self-comparison, and we obtain a code of $2W$ trinary digits.

\item The trinary code is transformed into a decimal number using Eq. \ref{eq:decimal}.
\begin{equation}
\small{
d = \sum _{j=0} ^ {2W} (t_j \times 3^j);} \hspace{0.3cm}
\end{equation}{\label{eq:decimal}}

We obtain $L-2W$ decimal numbers as the sliding window strides along the sequence of $L$ elements.

\item The $L-2W$ decimal values are collected into a histogram that summarises a patterns' statistic observed in the temporal sequence (without caring of the order). 

\end{enumerate}
At the end of this procedure, we obtain $K$ histograms describing the subsequent statistics of patterns. The quiet state is described by a histogram with a single central bin. The lowering of the central bin means that other bins are reaching non-zero configurations.
For this reason, we apply a bin threshold $t^*$ to the central value of the histogram, that characterises the absence of anomaly. We mark as anomaly the moment in which this histogram bin changes and triggers the threshold.  

The key observation of this process is that we model the normalcy in the scene through monitoring a trend of a specific extracted pattern over time. The histograms act as a summary of all the variations in the groups count, yet the bin that models the normalcy is only the one that represents a \textit{static} crowd dynamics (changes in the number of groups below the threshold $T$).

\section{Experimental results}
\subsection{Data set}
The algorithm has been tested using the Motion Emotion (ME) data set \cite{rabiee2016novel}. It contains 31 video sequences of around 44000 frames in total. Each video is recorded as 30 frames per second using a fixed video camera elevated at height, overlooking individual walkways and a video resolution of $554 \times 235$. The crowd density in the videos ranges from sparse to very crowded. The videos contain both normal and abnormal behavior, labeled frame-by-frame as 5 classes (Panic, Fight, Congestion, Obstacle and Neutral).
The videos do not capture only one behavior at a time, but rather the transition from normal to abnormal situation. For each behavior, at least two videos from different point of view are recorded. 
\begin{table}[]
\centering
\caption{Overview of the papers that use the ME dataset.}
\resizebox{0.49\textwidth}{!}{%
\begin{tabular}{|c|c|c|c|c|c|}
\hline
\multirow{2}{*}{\textbf{Refer.}}                                                 & \multirow{2}{*}{\textbf{Method}}                                         & \multicolumn{2}{c|}{\textbf{Application}}                                                                & \multicolumn{2}{c|}{\textbf{Performance}} \\ \cline{3-6} 
                                                                                    &    &\textbf{\begin{tabular}[c]{@{}c@{}}Anomaly \\ Detection\end{tabular}} & \textbf{\begin{tabular}[c]{@{}c@{}}Behavior \\ Classification\end{tabular}} & \textbf{Acc[\%]}     & \textbf{Avg Error}     \\ \hline
\multirow{5}{*}{\cite{rabiee2016novel}}                & Trajectory                                                               &                            & \checkmark                                                   & 35.30            & -                      \\ \cline{2-6} 
                                                                                    & HOG                                                                      &                            & \checkmark                                                   & 38.80            & -                      \\ \cline{2-6} 
                                                                                    & HOF                                                                      &                            & \checkmark                                                   & 37.69            & -                      \\ \cline{2-6} 
                                                                                    & MBH                                                                      &                            & \checkmark                                                   & 38.53            & -                      \\ \cline{2-6} 
                                                                                    & Dense Trajectory                                                         &                            & \checkmark                                                   & 38.71            & -                      \\ \hline
\multirow{2}{*}{\cite{varghese}} & Emotion Attribute                                                        &                            & \checkmark                                                   & 43.6             & -                      \\ \cline{2-6} 
                                                                                    & Spatio-Temporal                                                          &                            & \checkmark                                                   & 71.70            & -                      \\ \hline
 \cite{8553620}                                    & \begin{tabular}[c]{@{}c@{}}Density Heatmaps/\\ Optical Flow\end{tabular} &                            & \checkmark                                                   & 90.9             & -                      \\ \hline
\cite{briassouli}                                     & \begin{tabular}[c]{@{}c@{}}Phase-Based \\ Statistics\end{tabular}        & \checkmark  &                                                                             & -                & 0.118 \\ \hline
\end{tabular}}
\label{tab:comp}
\end{table}

Table \ref{tab:comp} shows the works that use the ME dataset in their experiments. Most of these papers focus on the behavior classification task. Even among them, it is difficult to compare the performances since the experiments do not have a common protocol. The only work on the detection anomaly task \cite{briassouli} analyzes the performance of a ``change detector'', considering not only the onset of the anomaly but also its resolution.
\subsection{Experimental protocol and evaluation parameters}
The anomaly detection performance depends on:
\begin{itemize}
\item the alarm synchronization with the anomaly occurrence - a maximum time delay can be taken into account in order to counteract appropriately;
\item ``low'' false positive detection rate - false alarms reduces the effectiveness of the system; for example, too many alarms over time are an obstacle during check of true anomaly occurrences.
\end{itemize}
To evaluate the reliability of the proposed method, we consider as anomalies correctly detected all those alarms that fall within a range corresponding to about 27 seconds, centred on the actual occurrence of the anomaly. This occurrence is manually labelled, by following the perception of the human operator.

If several alarms fall in this range, only one is considered. We considered the period before the occurrence of the anomaly because some changes in the group numbers may represent the dynamics leading to the anomaly.

\begin{figure}[ht]
\centering
\begin{subfigure}[b]{0.49\linewidth}
\centering\includegraphics[width=1\textwidth]{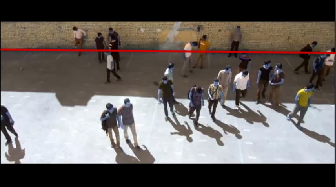}
\subcaption{}
\end{subfigure} 
\begin{subfigure}[b]{0.49\linewidth}
\centering\includegraphics[width=1\textwidth]{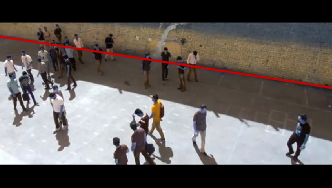}
\subcaption{}
\end{subfigure}
\caption{Videos partition in two separate sets by the camera position: the frontal videos have the line between the wall and the ground with a zero or small slope (a), the lateral videos have a visible slope (b). }
\label{fig:cameras}
\end{figure}

We divided the videos in two separate sets according to the camera position. In fact, some videos of the ME data set present a slope between wall and ground more visible than others ($> 5^\circ$). We have included these videos in the set of ``lateral camera'' and those with a very small slope in the set of ``frontal camera''. Fig. \ref{fig:cameras} shows an example of lateral and frontal view. These views reflect two real case scenarios.

We performed the following experiments:
\begin{itemize}
\item Experiment 1: set of videos recorded by the camera in frontal position with respect to the scene. The set includes videos [001, 002, 005, 006, 009, 010, 015, 016, 017, 018, 019, 020, 021, 022, 023, 024, 025, 026, 028, 029, 030, 031] of the Motion Emotion Dataset.
\item Experiment 2: set of videos recorded by the camera in lateral position with respect to the scene. The set includes videos [003, 004, 007, 008, 011, 012, 013, 014, 027] of the Motion Emotion Dataset.
\end{itemize}

We adopted the precision, recall and F1 score metrics to evaluate the overall system performance since anomalous detection is characterized by an uneven classes distribution: the number of frames with anomalies are considerably smaller than those without them.

We plotted the timeline description of the detection (Figs. \ref{fig:video9}-\ref{fig:video25}(b)): the anomaly labels are green vertical lines and the corresponding alarms are represented by red vertical lines. The 27-seconds window centred on the anomaly is highlighted in light green. Each alarm output into that time window is evaluated as a true positive (TP). Eqs. \ref{prec}-\ref{f1} define the precision, recall and F1 metrics, respectively:

\begin{equation}
\small{
    precision= \frac{TP}{TP + FP}}
    \label{prec}
\end{equation}
\begin{equation}
\small{
    recall= \frac{TP}{TP + FN}}
        \label{rec}
\end{equation}
\begin{equation}
\small{
    F1_{score}=2*\frac{precision*recall}{precision+recall}}
        \label{f1}
\end{equation}

Where $TP$ is the number of correctly detected anomalies, $FP$ is the number of false alarms, and $FN$ is the number of undetected anomalies.

Table \ref{tab:descriptor-params} shows the parameters shared in all the experiments. The remaining parameters have been set using a grid search, which maximize the $F1_{score}$. The ranges of the parameters set by grid search are shown in Tab. \ref{tab:grid-search}. The grid search was done both in a supervised way, i.e. the $F1_{score}$ is maximised on all videos, and with a Leave-one-out cross validation, i.e. the parameters maximize the $F1_{score}$ on $N-1$ videos and then they are used to test the video left out. Results are reported in Table \ref{tab:descriptor-params-exp-2} for each Experiment. Although supervised experiments lead to a biased estimate, they allow us to show the optimal performances obtainable with the perfect setting of the parameters.

\begin{table}[htpb!]
\centering
\caption{Common parameters used for all the experiment.}
\begin{tabular}{l r}
\hline
\multicolumn{2}{c}{\textbf{PARAMETERS EXPERIMENT}}\\
\hline
video frame rate & 30\\
skipped frames ($F$) & 20\\
trinary code sliding window size ($2W+1)$ & 5\\
\hline

\end{tabular}
\label{tab:descriptor-params}
\end{table}
\begin{table}[htpb!]
\centering
\caption{Tested parameters for tuning the algorithm.}
\begin{tabular}{l r}
\hline
\multicolumn{2}{c}{\textbf{GRID SEARCH RANGES}}\\
\hline
 descriptor threshold ($T$) & [2, 3, 4, 5, 6] \\
 descriptor sliding window size ($L$) & [10, 15, 20, 25, 30] \\
 bin threshold ($t^*$)  & [0.5, 0.55, 0.6, 0.65,\\ & 0.7, 0.75, 0.8, 0.85, 0.9, 0.95]\\
\hline
\end{tabular}
\label{tab:grid-search}
\end{table}

\begin{table}[htpb!]
\centering
\caption{Parameters used for the experiments 1 and 2. For the Leave-one-out (LOO), the average and the standard deviation of the parameters used for each video were reported.}
\resizebox{0.49\textwidth}{!}{%
\begin{tabular}{|c|c|c|c|c|c|c|c|}
\hline
\multicolumn{2}{|c|}{\multirow{2}{*}{\textbf{\begin{tabular}[c]{@{}c@{}}Grid search \\ parameters\end{tabular}}}} & \multicolumn{2}{c|}{\textbf{T}}                & \multicolumn{2}{c|}{\textbf{L}}                 & \multicolumn{2}{c|}{\textbf{t*}}               \\ \cline{3-8} 
\multicolumn{2}{|c|}{}                                                                                               & \textbf{Sup} & \textbf{LOO}                 & \textbf{Sup} & \textbf{LOO}                  & \textbf{Sup} & \textbf{LOO}                 \\ \hline
\multirow{2}{*}{\textbf{MC}}                                     & \textbf{Exp 1}                                    & $3$               & $4.25\pm1.09$ & $15$              & $15.00\pm0.00$ & $0.85$            & $0.76\pm0.15$ \\ \cline{2-8} 
                                                                 & \textbf{Exp 2}                                    & $2$               & $4.00\pm1.22$ & $20$              & $16.25\pm6.49$ & $0.65$            & $0.68\pm0.12$ \\ \hline
\multirow{2}{*}{\textbf{COF}}                                    & \textbf{Exp 1}                                    & $4$               & $3.50\pm0.50$ & $15$      & $15.00\pm0.00$ & $0.50$            & $0.68\pm0.17$ \\ \cline{2-8} 
                                                                 & \textbf{Exp 2}                                    & $6$               & $4.25\pm1.48$ & $10$              & $12.50\pm4.33$ & $0.85$            & $0.76\pm0.09$ \\ \hline
\multirow{2}{*}{\textbf{CD}}                                     & \textbf{Exp 1}                                    & $4$               & $4.25\pm1.09$ & $15$              & $15.00\pm0.00 $& $0.85$            & $0.76\pm0.15$ \\ \cline{2-8} 
                                                                 & \textbf{Exp 2}                                    & $4$               & $3.75\pm1.09$ & $10$              & $17.50\pm5.59$ & $0.85$            & $0.69\pm0.11$ \\ \hline
\multirow{2}{*}{\textbf{BD}}                                     & \textbf{Exp 1}                                    & $3$               & $3.75\pm0.83$ & $15$              & $18.75\pm6.50$ & $0.50$            & $0.74\pm0.17$ \\ \cline{2-8} 
                                                                 & \textbf{Exp 2}                                    & $5$               & $4.25\pm1.48$ & $10$              & $16.25\pm6.50$ & $0.70$            &$0.76\pm0.09$ \\ \hline
\end{tabular}
}
\label{tab:descriptor-params-exp-2}
\end{table}

\subsection{Overall Results}
We reported the performance measurements related to Experiment 1 and 2 in Tab. \ref{table:result_exp1} and \ref{table:result_exp2}.
Table \ref{table:result} illustrates  the overall performance results.
In each table, the performance refers to the proposed descriptor receiving as input the number of groups obtained by the counting methods described in Section \ref{sec:lowlevfeat}, namely, Clustering of Optical Flow (\textbf{COF}), Cascade Detector (\textbf{CD}) and Blob Detection (\textbf{BD}).
\begin{table*}[!ht]
\centering
\caption{Performance comparison of the proposed method using different counting methods for videos  recorded  by  the  camera in frontal  position.}
\resizebox{0.5\textwidth}{!}{%
\begin{tabular}{c|c|c|c|c|c|c|}
\cline{2-7}
                                            & \multicolumn{3}{c|}{\textbf{Supervised}}           & \multicolumn{3}{c|}{\textbf{Leave-one-out}}         \\ \hline
\multicolumn{1}{|c|}{\textbf{Experiment 1}} & \textbf{Precision} & \textbf{Recall} & \textbf{F1} & \textbf{Precision} & \textbf{Recall} & \textbf{F1}  \\ \hline
\multicolumn{1}{|c|}{\textbf{MC}}           & 84.61\%             &  91.67\%          & 88.00\%      & 68.42\%       & 59.09\%    & 63.41\% \\ \hline
\multicolumn{1}{|c|}{\textbf{COF}}          & 62.50\%             & 83.33\%          & 71.43\%      & 48.28\%       & 58.33\%    & 52.83\% \\ \hline
\multicolumn{1}{|c|}{\textbf{CD}}           & 70.97\%             & 91.67\%          & 80.00\%      &  70.97\%       & 91.67\%    & 80.00\% \\ \hline
\multicolumn{1}{|c|}{\textbf{BD}}           & 63.33\%             &  79.17\%          & 70.37\%      & 50.00\%       & 66.67\%    & 57.14\% \\ \hline
\end{tabular}}
\label{table:result_exp1}
\end{table*}

\begin{table*}[!ht]
\centering
\caption{Performance comparison of the proposed method using different counting methods for videos  recorded  by  the  camera in lateral  position.}
\resizebox{0.5\textwidth}{!}{%
\begin{tabular}{c|c|c|c|c|c|c|}
\cline{2-7}
                                            & \multicolumn{3}{c|}{\textbf{Supervised}}           & \multicolumn{3}{c|}{\textbf{Leave-one-out}}         \\ \hline
\multicolumn{1}{|c|}{\textbf{Experiment 2}} & \textbf{Precision} & \textbf{Recall} & \textbf{F1} & \textbf{Precision} & \textbf{Recall} & \textbf{F1}  \\ \hline
\multicolumn{1}{|c|}{\textbf{MC}}           & 100.00\%            & 100.00\%         & 100.00\%     & 100.00\%       & 100.00\%    & 100.00\% \\ \hline
\multicolumn{1}{|c|}{\textbf{COF}}          & 92.31\%             & 100.00\%         & 96.00\%      & 63.64\%       & 63.64 \%    & 63.64\% \\ \hline
\multicolumn{1}{|c|}{\textbf{CD}}           & 84.61\%             & 91.67\%          & 88.00\%      & 80.00\%       &66.67\%    & 72.73\% \\ \hline
\multicolumn{1}{|c|}{\textbf{BD}}           & 85.71\%            & 100.00\%          & 92.31\%      & 71.43\%       & 90.91\%    & 80.00\% \\ \hline
\end{tabular}}
\label{table:result_exp2}
\end{table*}

\begin{table*}[!ht]
\centering
\caption{Performance comparison of the proposed method using different counting methods for all ME data set videos.}
\resizebox{0.5\textwidth}{!}{%
\begin{tabular}{c|c|c|c|c|c|c|}
\cline{2-7}
                                             & \multicolumn{3}{c|}{\textbf{Supervised}}           & \multicolumn{3}{c|}{\textbf{Leave-one-out}}                                \\ \hline
\multicolumn{1}{|c|}{\textbf{All ME videos}} & \textbf{Precision} & \textbf{Recall} & \textbf{F1} & \textbf{Precision}           & \textbf{Recall}              & \textbf{F1}  \\ \hline
\multicolumn{1}{|c|}{\textbf{MC}}            & 88.89\%             & 94.12\%          & 91.43\%      & 79.31\% &  71.87\% & 75.41\% \\ \hline
\multicolumn{1}{|c|}{\textbf{COF}}           & 71.11\%             & 88.89\%          & 79.01\%      & 52.50\%                 & 60.00\%                 &  56.00\% \\ \hline
\multicolumn{1}{|c|}{\textbf{CD}}            & 75.00\%            & 91.67\%          & 82.50\%      & 73.17\%                 & 83.33\%                 & 77.92\% \\ \hline
\multicolumn{1}{|c|}{\textbf{BD}}            & 70.45\%             & 86.11\%          & 77.50\%      &  56.52\%                 & 74.29\%                 & 64.20\% \\ \hline
\end{tabular}}
\label{table:result}
\end{table*}
According to Experiment 1 and 2, the different position of the camera leads to very different performances. Lateral views appear to have a deeper observation surface than frontals (Figs. \ref{fig:cameras}(a-b)), and this might suggest that the descriptor has more ``time'' to observe changes in the scene, especially in case of anomalous events.

The group counting method affects the performance of the descriptor. The most reliable method is the Cascade Detector which on all videos achieves better detection performance than even the Manual Counting in the Leave-one-out protocol.

The difference in performance between the supervised and the leave-one-out protocol suggests that a more accurate setting of the parameters calculated by grid search would allow for more reliable detection. Another explanation is that the number of videos is too small for achieving a reliable estimation by cross-validation approaches.  

\subsection{Analysis of individual cases}
We analyze some videos in detail in order to explain how the proposed descriptor behaves.
In the following figures, the green lines represent the actual anomalies and the red lines the anomalies predicted by the system. If the red line lies in a light green area, the anomaly is considered correctly detected.

\begin{figure}[ht]
\centering
\begin{subfigure}[b]{0.5\linewidth}
\centering
\includegraphics[width=1\textwidth]{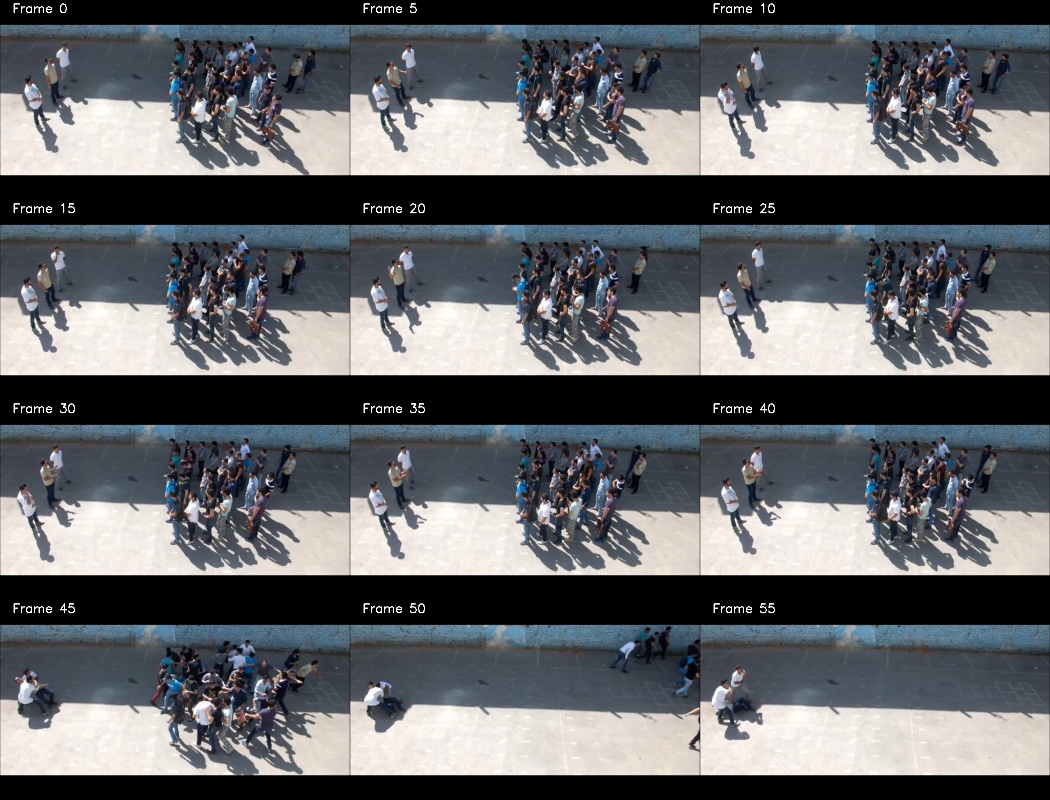}
\end{subfigure}\\
\begin{subfigure}[b]{0.5\linewidth}
\centering
\includegraphics[width=1\textwidth]{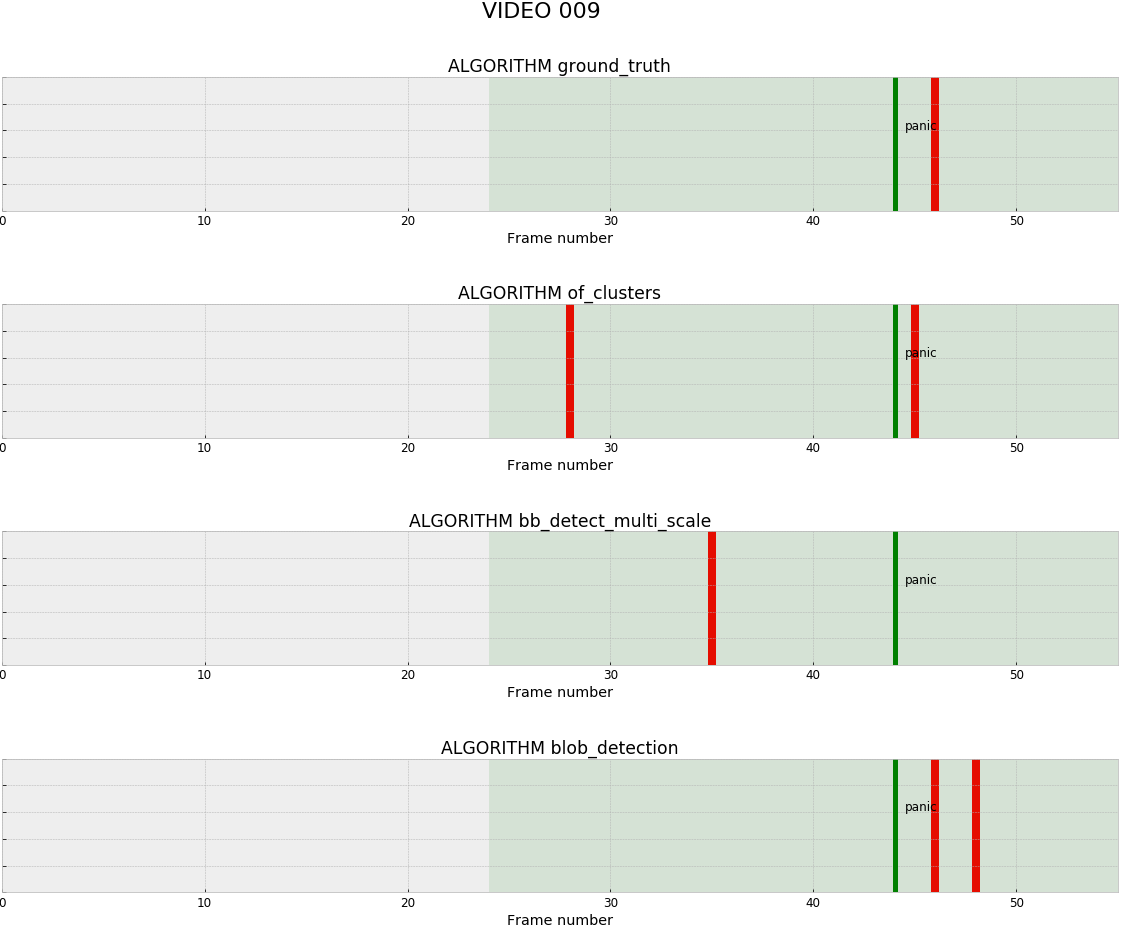}
\end{subfigure}
\caption{Frames summary and graphical description of detection for video 009: the video presents a panic situation that has been correctly identified by all the methods. }
\label{fig:video9}
\end{figure}

\begin{figure}[!ht]
\centering
\begin{subfigure}[b]{0.5\linewidth}
\centering
\includegraphics[width=1\textwidth]{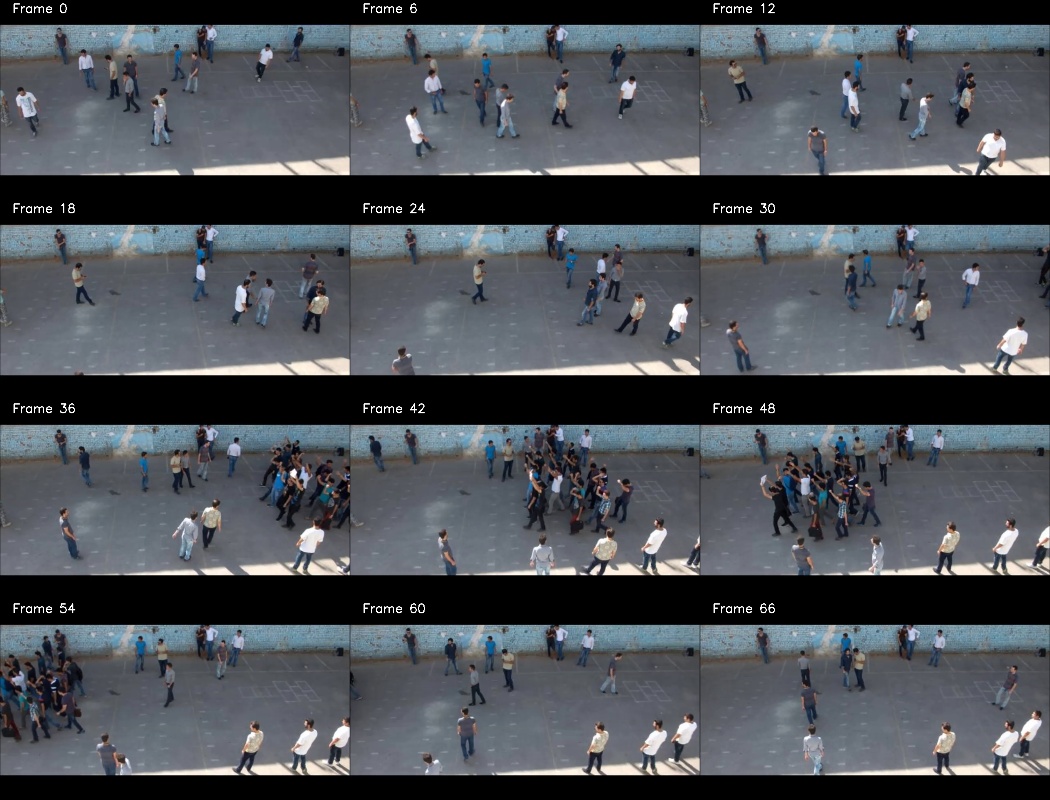}
\end{subfigure}\\
\begin{subfigure}[b]{0.5\linewidth}
\centering
\includegraphics[width=1\textwidth]{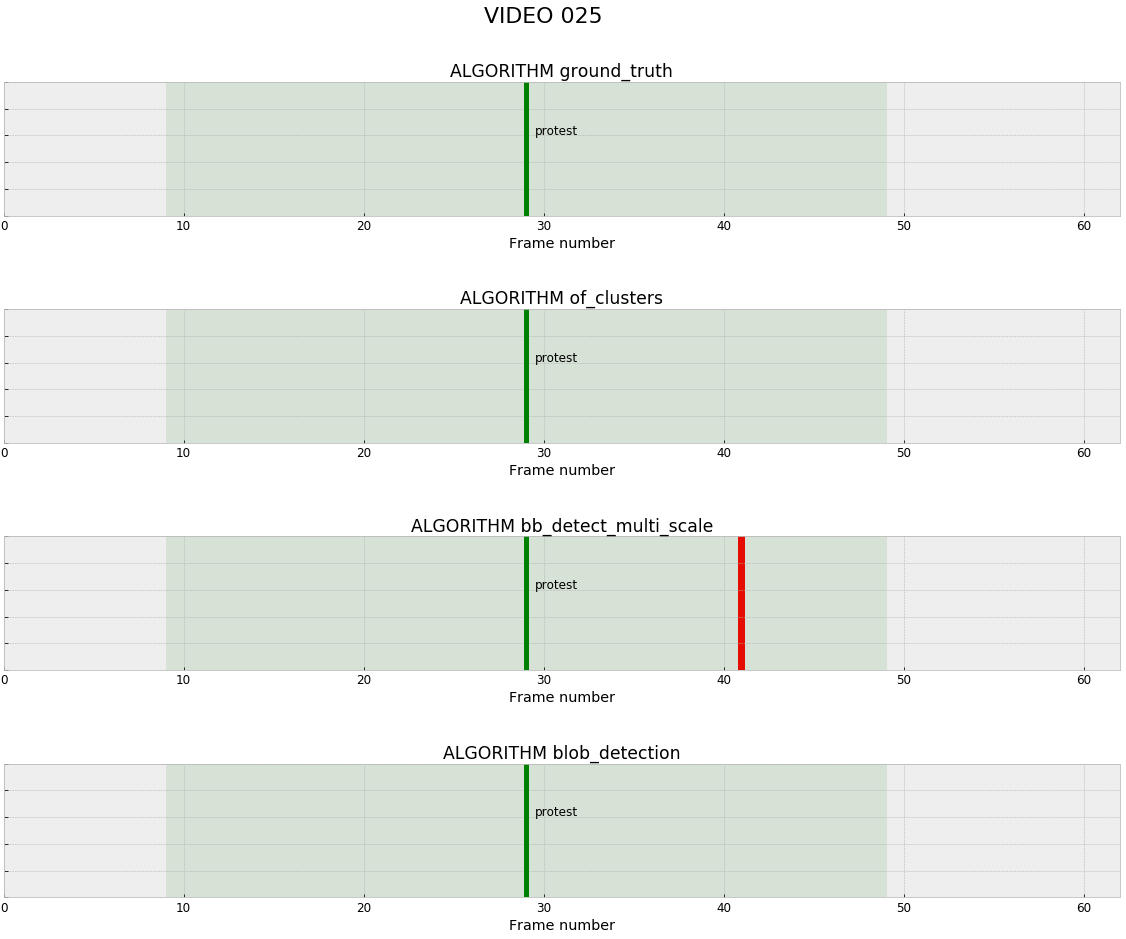}
\end{subfigure}
\caption{Frames summary and graphical description of detection for video 025: in the video a protest crosses the scene. This anamoly is difficult to identify because crowd change is slow and controlled.}
\label{fig:video25}
\end{figure}

\begin{figure}[!ht]
\centering
\begin{subfigure}[b]{0.5\linewidth}
\centering
\includegraphics[width=1\textwidth]{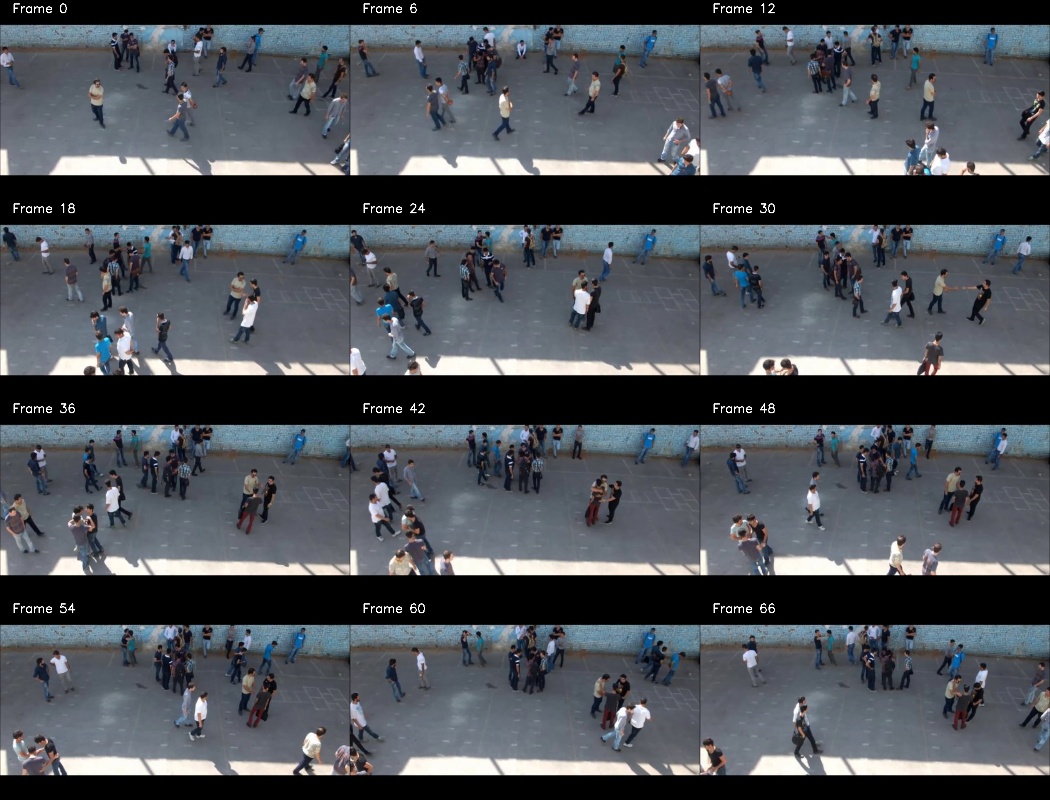}
\subcaption{}
\end{subfigure}\\
\begin{subfigure}[b]{0.5\linewidth}
\centering
\includegraphics[width=1\textwidth]{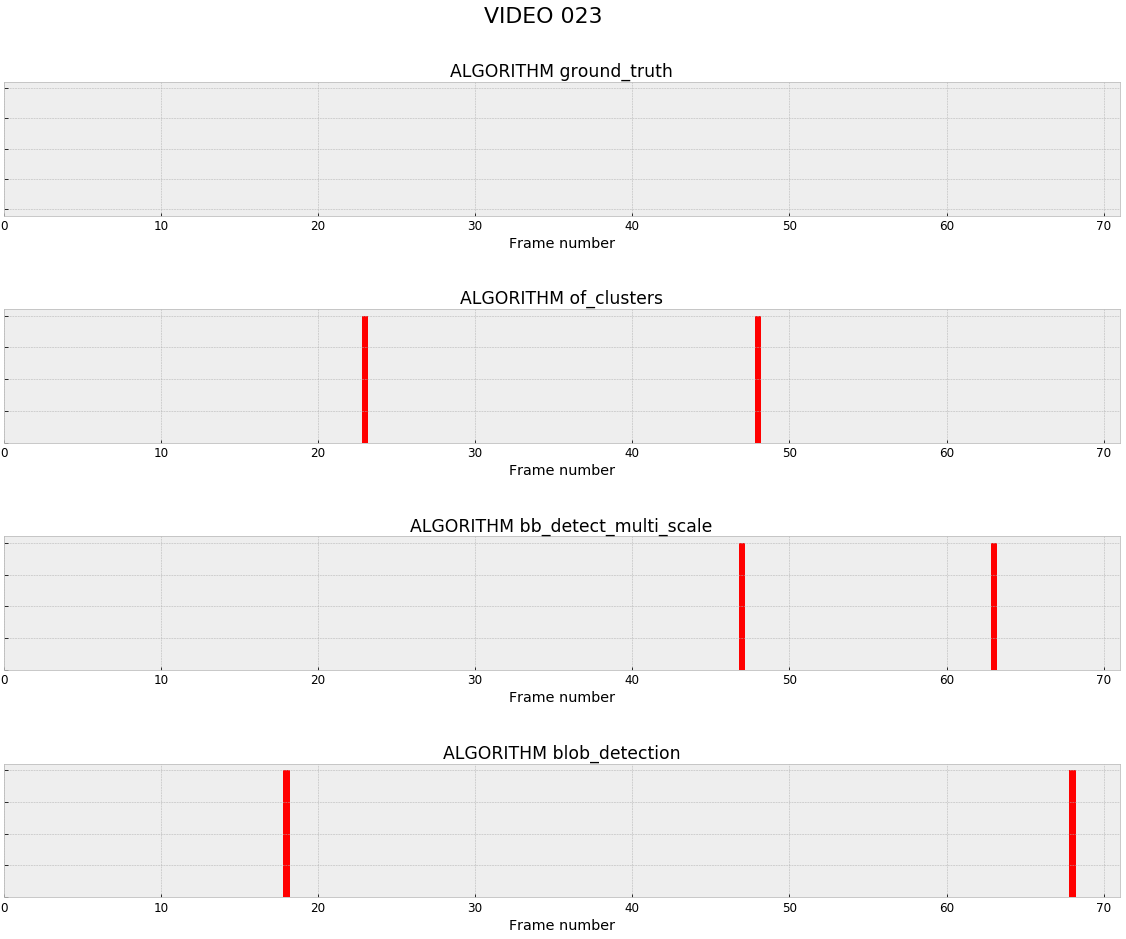}
\subcaption{}
\end{subfigure}
\caption{Frames summary and graphical description of detection for video 023: the video does not present anomalies but the descriptor leads to false alarms.}
\label{fig:video23}
\end{figure}

Firstly, we  selected the video $009$ to represent the case where the detector is able to correctly identify the anomaly with all the counting methods (Fig. \ref{fig:video9}).
The video is characterised by an initial static flow of individuals, namely, a structured crowd, and consequently a constant number of groups. The panic event generates movement whereby the detectors are able to notice the anomaly. This is fully in agreement with the hypothesis behind our work.
The video $025$ (Fig. \ref{fig:video25}) presents an anomaly (protest) that is not detected except by counting with Cascade Detector. This is due to the fact that the anomaly is characterized by a single group of people who cross the scene with a controlled and slow movement. The number of groups therefore has no particular variations. This is still a structured scene, and the behaviour of the detector is in agreement with our hypothesis.
The video $023$ (Fig. \ref{fig:video23}) does not present anomalies. However, this case is characterised by a large number of false alarms.
This is a case of unstructured and sparse crowd where we have very small changes in the number of groups. This behaviour is detected by the system and interpreted as multiple anomalies. This effect can be reduced by avoiding the bins of the trit-based histogram that represent such small variations\footnote{The histograms that describe the frames were not reported in the paper for reasons of space but can be seen as supplementary material.}.

\subsection{Computational time analysis}
The system configuration for determining the processing times of a frame was a MacBookPro with macOS Sierra, 2.7 GHz Intel Core i5, 8 GB 1867 MHz DDR3 of RAM and a Intel Iris Graphics 6100 1536 MB.
The total processing time is highly influenced by the group counting method, whose times are of the order of magnitude of hundreds of milliseconds and are reported in Tab. \ref{tab:computation-time}. The creation of the high-level features from the count values is in fact extremely fast, about 1 ms using a sliding window of size L=15.
Considering that we analyze one frame each F = 20 frames, namely one frame per 666 ms, we can build a pipeline without delay for all the methods analyzed. 
However, it should be noted that the cascade detector and the blob detector, in addition to being more performing, are faster than the clustering of optical flow.
\begin{table}[htpb!]
\centering
\caption{Time for the processing of a single frame with the implemented algorithms.}
\resizebox{0.17\textwidth}{!}{%
\begin{tabular}{l r}
\hline
\textbf{Algorithm} & \textbf{ms per frame}\\
\hline
COF & $\sim 650$ \\
CD & $\sim 150$ \\
BD & $\sim 100$ \\
\hline
\end{tabular}}
\label{tab:computation-time}
\end{table}

\section{Concluding remarks}
In this work, we proposed a novel temporal descriptor of small and large crowds by computer vision algorithms. We exploited the typical rapid aggregation and disruption of crowds during anomalous events and the concept of one-dimensional binary pattern. The former allowed us to focus on the number of groups of individuals over time, the latter on its variations, leading to describe them by what we called trit-based bins. 

We used the Motion-Emotion data set to benchmark the proposed system and three different methods for group counting. 
As expected, the descriptor works well with events that clearly match our hypothesis, that is, the number of groups changes quickly and presents errors if the disruptions and aggregations are slow and controlled. On the other hand, further experiments must be aimed at analysing the trit-based bins to reduce the number of false alarms that may appear in case of sparse and unstructured crowds.

Although the performance is not good enough for real applications, the advantage of being a white box makes the descriptor easily adaptable depending on the real context and the type of anomaly to be detected.

\section*{Acknowledgment}
This work is supported by the Italian Ministry of Education, University and 
Research (MIUR) within the PRIN2017 - BullyBuster - A framework for bullying and cyberbullying action detection by computer vision and artificial intelligence 
methods and algorithms (CUP: F74I19000370001).




\bibliographystyle{abbrv}
\bibliography{bib}
%

\end{document}